\documentclass[conference]{IEEEtran}
\IEEEoverridecommandlockouts
\usepackage{cite}
\usepackage{amsmath,amssymb,amsfonts}
\usepackage{algorithmic}
\usepackage{graphicx}
\usepackage{textcomp}
\usepackage{bm}

\usepackage{multirow}
\usepackage[table,xcdraw]{xcolor}
\def\BibTeX{{\rm B\kern-.05em{\sc i\kern-.025em b}\kern-.08em
    T\kern-.1667em\lower.7ex\hbox{E}\kern-.125emX}}

\begin{document}
\title{Edge-InversionNet: Enabling Efficient Inference of InversionNet on Edge Devices\\
\thanks{* This work is under review for publication in conjunction with IMAGE ‘23}
}

\author{\IEEEauthorblockN{
Zhepeng Wang\textsuperscript{\dag, \S},
Isaacshubhanand Putla\textsuperscript{\dag}, 
Weiwen Jiang\textsuperscript{\dag}, 
Youzuo Lin\textsuperscript{\S}}

\IEEEauthorblockA{\textsuperscript{\dag}George Mason University, Department of Electrical and Computer Engineering, VA, USA.\\
\textsuperscript{\S}Los Alamos National Laboratory, Energy and Natural Resources Security, NM, USA.\\
\vspace{-0.15in}}
}

\maketitle

\begin{abstract}
Seismic full waveform inversion (FWI) is a widely used technique in geophysics for inferring subsurface structures from seismic data. And InversionNet is one of the most successful data-driven machine learning models that is applied to seismic FWI. However, the high computing costs to run InversionNet have made it challenging to be efficiently deployed on edge devices that are usually resource-constrained. Therefore, we propose to employ the structured pruning algorithm to get a lightweight version of InversionNet, which can make an efficient inference on edge devices. And we also made a prototype with Raspberry Pi to run the lightweight InversionNet. Experimental results show that the pruned InversionNet can achieve up to 98.2 \% reduction in the computing resources with moderate model performance degradation.

\end{abstract}


\section{Introduction}

Seismic full waveform inversion (FWI) is  extensively employed in identifying important geophysical properties such as velocity and conductivity. And it is thus widely used in diverse
subsurface applications including subsurface energy exploration, earthquake early warning systems, and carbon capture and sequestration, etc~\cite{virieux2009overview, lin2023physics, feng2022extremely, wu2018deepdetect}. 

To solve the seismic full waveform inversion problem, two types of methods are proposed, which are the physics-driven method~\cite{lin2015quantifying, feng2021multiscale} and the data-driven method~\cite{yang2019deep, adler2020deep, feng2022intriguing}. The physics-driven method can obtain the subsurface characteristics using governing physics and equations, without strong dependence on data, while the data-driven method utilizes machine learning models to learn the relationship between the seismic measurements and the subsurface structure from a large volume of training data.

\begin{figure}[ht]
\centering
\includegraphics[width=1\columnwidth]{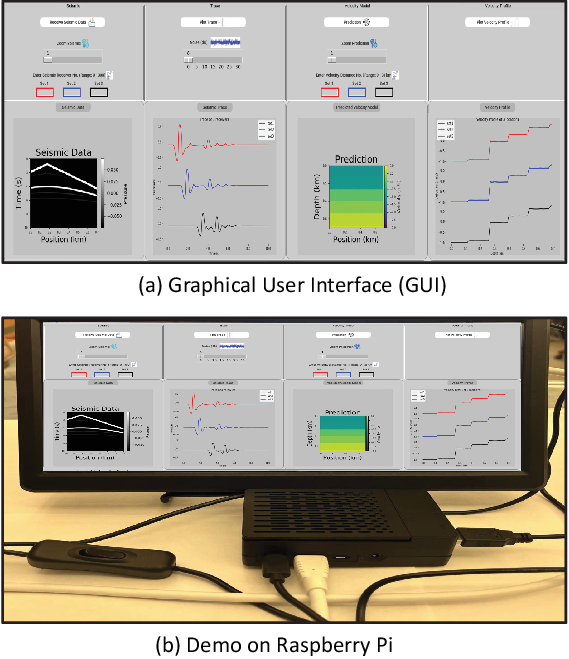}
\caption{Prototype of Edge-InversionNet: (a). Graphical User Interface (GUI). (b) Demo on Raspberry Pi.}
\label{fig:prototype}
\end{figure}

Compared with the physics-driven method, the data-driven method can usually generate a higher resolution for the detection of small structures in a more computationally efficient way, owing to the high expressive power of the machine learning model. InversionNet~\cite{wu2019inversionnet} is one of the most successful models in data-driven methods. It can consistently achieve the SOTA performance on a wide range of benchmark datasets and is also a fundamental model with many variants~\cite{zhang2020data, zeng2021inversionnet3d, jin2022unsupervised, manu2022seismic}.

However, making inferences with InversionNet requires high computing resources, which impedes it from running on resource-constrained edge devices. Without efficient deployment onto edge devices, it is challenging to apply InversionNet to the scenario where the on-device processing of the data near the source of its acquisition is important due to the high requirements for data privacy and real-time decision-making.

\begin{figure*}[ht]
\centering
\includegraphics[width=2\columnwidth]{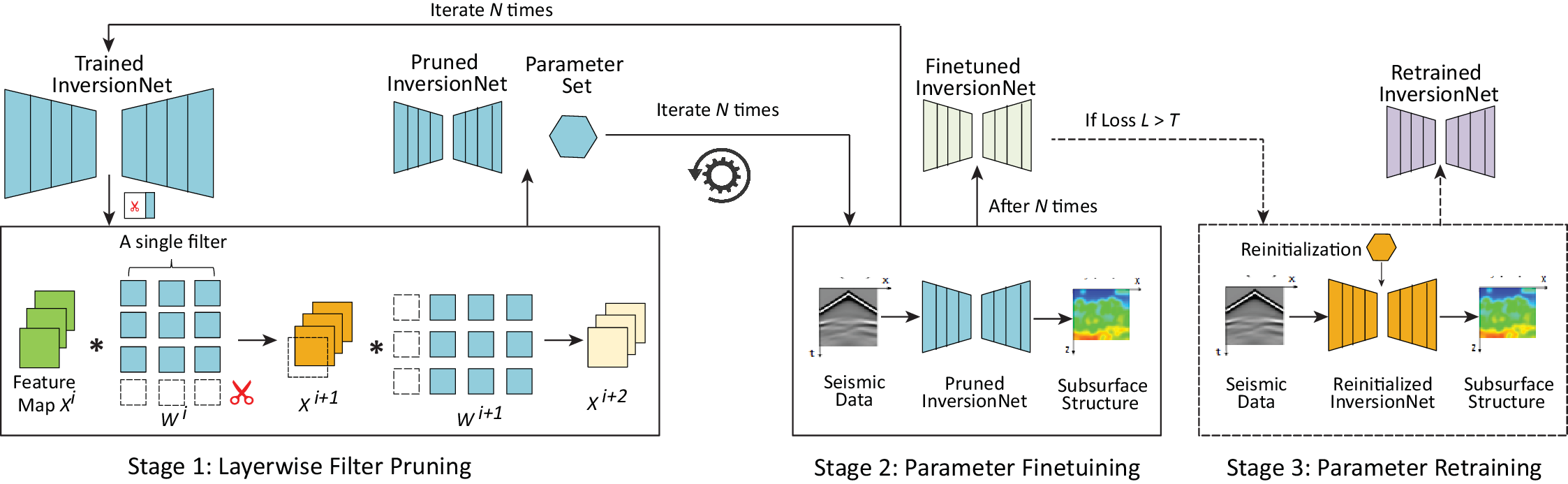}
\caption{Overview of the pruning algorithm in Edge-InversionNet}
\label{fig:Overview}
\end{figure*}
To enable efficient inference of InversionNet on edge devices, we propose to utilize structured pruning~\cite{li2017pruning,liu2019metapruning,li2022revisiting} to generate a lightweight version of InversionNet. More specifically, given a pre-trained InversionNet, the pruning algorithm can identify and remove the filters that have a minor effect on the model performance in a progressive way, such that effectively reducing the size and the computing costs of the model without hurting the model performance significantly. The parameters of the pruned InversionNet will then be fine-tuned or retrained to further recover the model performance. Compared with unstructured pruning~\cite{han2015deep_compression, peng2022towards, wang2020sparsert, bao2020fast,bao2022accelerated, bao2022doubly}, structured pruning is hardware-friendly, which means the generated lightweight model can be directly deployed to off-the-shelf edge devices without customized hardware support.

Although structured pruning has made great achievements in the machine learning model for image processing, to the best of our knowledge, there are few works employing it in the machine learning model designed for the seismic FWI. Our work demonstrates the potential of structured pruning to generate lightweight models to solve FWI problems efficiently on edge devices. Moreover, we developed a prototype of a system called Edge-InversionNet, which is shown in Fig.~\ref{fig:prototype}. The prototype can run the pruned InversionNet for inference on Raspberry Pi in real time and visualize the input and output with the help of a graphical user interface (GUI). The experimental results show that the pruned InversionNet can have about $73.0\%$ reduction in the computing resources with negligible performance degradation and $ 98.2\%$ reduction in the computing resources with moderate loss of performance. 


\section{Method}
\subsection{Overview}
\begin{figure*}[!t]
\centering
\includegraphics[width=2\columnwidth]{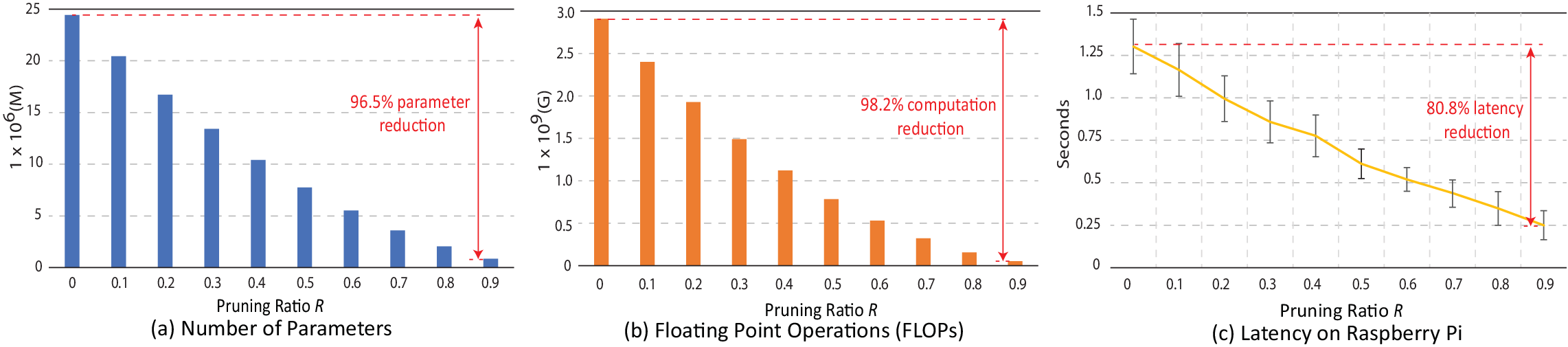}
\caption{Evaluation of the computing costs: (a) Number of Parameters, (b) Floating Point Operations (FLOPs) and (c) Latency on Raspberry Pi}
\label{fig:hardware}
\end{figure*}
Here we apply structured pruning~\cite{li2017pruning,liu2019metapruning,li2022revisiting} to a trained InversionNet $M$. More specifically, given a pruning ratio $R$, the number of fine-tuning iterations $N$, and the loss threshold $T$, the pruning algorithm prunes the filters at each layer of $M$ at a ratio of $R$ uniformly (i.e., only $1-R$ filters are preserved) and aims to maintain the performance of the lightweight InversionNet $M^{\prime}$, where we have the loss $L(M^{\prime})\leq T$.

Fig.~\ref{fig:Overview} shows the overview of the pruning algorithm. There are at most three stages. Stage 1 first prunes the filters of a trained InversionNet $M$ at each layer at a ratio $r$, which is equal to $1- \sqrt[N]{1-R}$. Then the pruned InversionNet will be sent to stage 2 with its reduced parameter set $\bm{W}$. In stage 2, the pruned InversionNet initialized with $\bm{W}$ is finetuned on the training set to recover its performance. Then the finetuned InversionNet will be sent back to stage 1 to serve as the trained InversionNet. This process iterates for $N$ times, such that the final pruning ratio of the model can achieve $R$. Then the finetuned InversionNet is evaluated on the validation set to get the loss $L$. If $L\leq T$, the finetuned InversionNet will be the output lightweight InversionNet $M^{\prime}$ of the pruning algorithm. Otherwise, the finetuned InversionNet is sent to stage 3 and its parameter set is reinitialized randomly. Then the reinitialized InversionNet is retrained on the training set from scratch and evaluated on the validation set. Comparing the loss of finetuned InversionNet and the retrained InversionNet, the one with a smaller loss is selected to be the lightweight InversionNet $M^{\prime}$ generated by the pruning algorithm.

\subsection{Stage 1: Layerwise Filter Pruning} 
In each iteration, given a pruning ratio $r$, the pruning algorithm needs to find the best strategy to reduce the number of filters in layer $i$ from $n_i$ to $n_i^{\prime}$, where $r = \frac{n_i-n_i^{\prime}}{n_i} (i = 1, 2, .., K)$ and $K$ is the total number of layers in model $M$. To minimize the loss $L$ of the pruned model, the problem can be formulated as follows.
\begin{equation}~\label{pruning}
\begin{split}
& \underset{\boldsymbol{\beta}}{\arg \min } \quad L(\boldsymbol{W}, \boldsymbol{\beta}) \\
& \text{s.t.}\quad \|\boldsymbol{\beta_{i}}\|_{0} = n_i^{\prime},\quad \forall \boldsymbol{\beta_{i}} \in \boldsymbol{\beta},\ i = 1, 2, .., K
\end{split}
\end{equation}
Where $\boldsymbol{W}$ is the parameter set of model $M$ and $\boldsymbol{\beta}$ represents the filter selection strategy produced by the pruning algorithm. More specifically, ${\boldsymbol{\beta_{i}}}$ is the strategy for layer $i$, which is represented as a binary vector with $n_i$ dimension. And if the $j$th element of ${\boldsymbol{\beta_{i}}}$ is $0$, the $j$th filter in layer $i$ will be pruned.

Directly solving the optimization problem in equation~\ref{pruning} with exhaustive search is computationally prohibitive since the total number of possible cases is  $\prod_{i=1}^{K}\tbinom{n_i}{r\cdot n_i}$ and for each case, we need to run the corresponding pruned model on the whole training set to get the loss. 

Therefore, we utilize $\ell_{1}$-norm of the weights within the filter to serve as a proxy metric to quantify the importance score of the filter. And its  computation cost is small. More specifically, for each filter $j$ in layer $i$, its importance score $s_{i}^{j}$ is $\left\|\boldsymbol{W}_{i}^{j}\right\|$, where $\boldsymbol{W}_{i}^{j}$ corresponds to the parameters of filter $j$ in layer $i$. After calculating the important scores of all the filters, we will prune the $r\cdot n_i$  filters with the smallest scores in layer $i$.

As shown in Fig.~\ref{fig:Overview}, if we prune one filter in layer $i$, the corresponding channel in the output $X^{i+1}$ will also be removed. And it can cause the removal of one kernel for each filter in layer $i+1$ even if the filter pruning has not been applied to layer $i+1$ yet. Therefore, the size of the pruned model will be reduced by $R^2$ when the given pruning ratio of the model is $R$. And this conclusion can also be verified by our experimental results.

\subsection{Stage 2: Parameter Finetuning} 
The pruned model from stage 2 usually suffers from performance degradation due to the reduction of parameters. But the value of the remaining parameter set is usually a good initialization that further finetuning can start with. By fintuning on the training set, the performance of the model on the validation set can usually be recovered. If the loss $L \leq T$, then the finetuned model will be the output of the pruning algorithm.

\subsection{Stage 3: Parameter Retraining} 
This stage is optional and is only activated when $L > T$. It indicates that the value of the remaining parameter set is misleading and makes the finetuning converge to a poor local minimum. Such a case usually occurs when the pruning ratio $R$ is quite large like $0.9$. Therefore, to mitigate performance degradation, extra weight retraining is introduced. More specifically, the selection of the filter still inherits from the pruning in stage 1 but the value of the remaining parameter set is dropped off. The parameter set is reinitialized randomly and retrained on the training set. The retrained model will then be evaluated on the validation set. If the loss of the retrained model is smaller than that of the finetuned model, it will be the output of the pruning algorithm. Otherwise, the finetuned model will be the output.
\begin{table*}[!ht]
\centering
\caption{Evaluation of the performance of the pruned InversionNet on OpenFWI}
\label{tab:perf_eval}
\setlength\tabcolsep{11 pt}
\begin{tabular}{c|ccc|ccc|ccc}
\hline
 & \multicolumn{3}{c|}{No Pruning (Baseline)} & \multicolumn{3}{c|}{Pruning Ratio $R$ (0.5)} & \multicolumn{3}{c}{Pruning Ratio $R$ (0.9)} \\ \cline{2-10} 
\multirow{-2}{*}{Dataset} & MAE $\downarrow$ & RMSE $\downarrow$ & SSIM $\uparrow$ & MAE $\downarrow$ & RMSE $\downarrow$ & SSIM $\uparrow$ & MAE $\downarrow$ & RMSE $\downarrow$ & SSIM $\uparrow$ \\ \hline
FlatVel-A & 0.0114 & 0.0193 & 0.9903 & 0.0155 & 0.0231 & 0.9894 & 0.0356 & 0.0618 & 0.9058 \\
FlatVel-B & 0.0347 & 0.0873 & 0.9482 & 0.0411 & 0.1011 & 0.9333 & 0.0864 & 0.1676 & 0.8227 \\
CurveVel-A & 0.0648 & 0.1233 & 0.8170 & 0.0707 & 0.1286 & 0.8134 & 0.1014 & 0.1632 & 0.7462 \\
CurveVel-B & 0.1496 & 0.2877 & 0.6742 & 0.1701 & 0.3083 & 0.6424 & 0.2622 & 0.4075 & 0.5152 \\
FlatFault-A & 0.0185 & 0.0436 & 0.9766 & 0.0222 & 0.0547 & 0.9644 & 0.0410 & 0.0945 & 0.9173 \\
FlatFault-B & 0.1044 & 0.1708 & 0.7267 & 0.1178 & 0.1827 & 0.7157 & 0.1547 & 0.2243 & 0.6589 \\
CurveFault-A & {\color[HTML]{000000} 0.0268} & 0.0674 & {\color[HTML]{000000} 0.9526} & 0.0331 & 0.0833 & 0.9367 & 0.0544 & 0.1202 & 0.8829 \\
CurveFault-B & 0.1604 & 0.2400 & 0.6214 & 0.1794 & 0.2601 & 0.5960 & 0.2112 & 0.2952 & 0.5497 \\
Style-A & 0.0618 & 0.1015 & 0.8874 & 0.0719 & 0.1130 & 0.8692 & 0.0910 & 0.1375 & 0.8208 \\
Style-B & 0.0582 & 0.0934 & 0.7515 & 0.0629 & 0.1000 & 0.7279 & 0.0827 & 0.1186 & 0.6583 \\ \hline
\end{tabular}
\end{table*}


\section{Experimental results}
\begin{figure*}[!t]
\centering
\includegraphics[width=2\columnwidth]{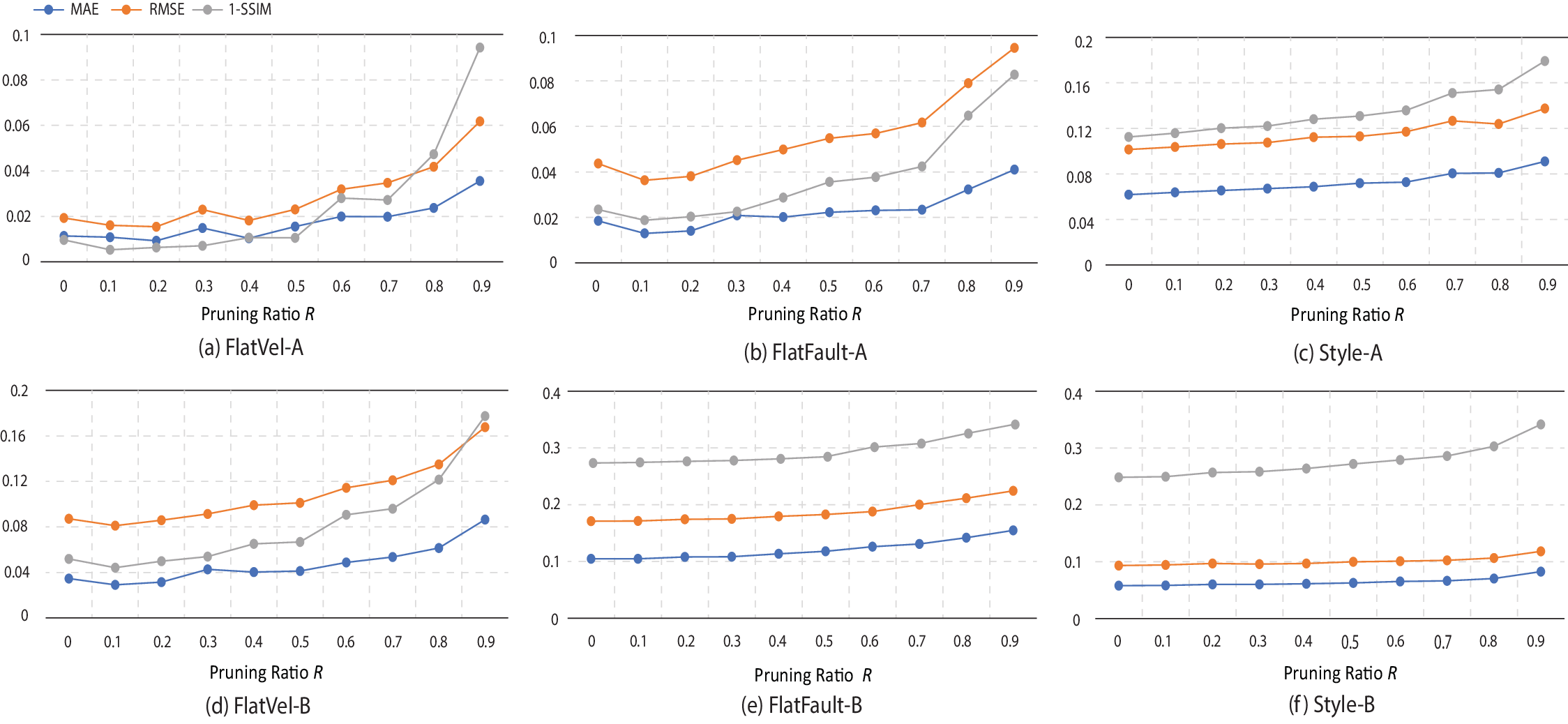}
\caption{Evaluation of performance on 6 selected datasets in OpenFWI}
\label{fig:performance}
\end{figure*}
In this section, we evaluate the structured pruning algorithm of Edge-InversionNet with different pruning ratio $R$ on OpenFWI, a recent and open-source benchmark datasets for seismic full waveform inversion~\cite{deng2022openfwi}. We employ 10 datasets in OpenFWI, including `Vel' family (30K samples), `Fault' family (54K samples) and `Style' family (67K samples). The size of the input is $5 \times 1000 \times 1 \times 70$, which corresponds to number of sources $\times$ number of time steps $\times$ number of receivers in width $\times$ number of receivers in length. And the output size is $70 \times 1 \times 70$, which corresponds to depth $\times$ width $\times$ length of the velocity map.

\subsection{Evaluation of the Computing Costs}
Fig.~\ref{fig:hardware} shows the evaluation of the computing costs of InversionNet pruned with different pruning ratio $R$. As shown in Fig.~\ref{fig:hardware} (a), the number of parameters of the model is decreased near-quadratically when the pruning ratio $R$ is increased. The reason why the trend is not exactly quadratic is that the necessary rounding operation to make the number of pruned filters an integer. Compared with the original InversionNet, $96.5 \%$ of parameters is pruned at the ratio of $0.9$, making the pruned model occupy much smaller space in memory and storage on the edge device. And the saved space  can then be used for saving and loading a larger batch of data on the device. 

To measure the amount of computation to run InversionNet, we calculate the total floating point operations (FLOPs) for a single inference of InversionNet. As shown in Fig.~\ref{fig:hardware} (b), FLOPs of the pruned InversionNet are also reduced near-quadratically along with the growth of the pruning ratio $R$. More specifically, $98.2\%$ computation can be saved when $R$ is $0.9$. Such a huge reduction can lower the energy consumption of inference, which extend the lifespan of the edge device without a battery replacement.

We also execute the inference of InversionNet pruned by different $R$ on the Raspberry Pi with 50 runs. The average latency and the standard deviation are plotted in Fig.~\ref{fig:hardware} (c). $80.8\%$ of time is saved to run an inference when $R$ is $0.9$, which takes the runtime overhead of the Raspberry Pi into consideration. The latency reduction is conducive to achieving real-time decision-making on the edge device. 

\subsection{Evaluation of the Performance}
To show the performance of the pruned InversionNet, we run the pruning algorithm in Edge-InversioNet at the pruning ratio of $R$ with $N$ iterations, where $R \in \{0.1, 0.2, .., 0.9\}$ and $N \in \{1, 3, 5\}$. The loss function for parameter finetuning and retraining is mean absolute error (MAE). And the total epoch for fintuning and retraining is both 120 epochs. Note that the number of epochs for each iteration during finetuning is $\frac{120}{N}$.

In Table~\ref{tab:perf_eval}, we report the best performance of the pruned InversionNet when $R$ is $0.5$ and $0.9$. The performance of the InversionNet without pruning is served as the baseline. Here we use three metrics for evaluation: mean absolute error (MAE), rooted mean squared error (RMSE) and structural similarity (SSIM). MAE and RMSE quantify the numerical disparity between the predicted and true velocity maps, while SSIM captures the perceptual similarity of the two maps. More specifically, when $R$ is $0.5$, the performance of the model is consistently close to that of the baseline on all the evaluated datasets in terms of all the metrics, which only incurs a minor loss of performance. The performance degradation becomes more obvious when the pruning is aggressive, i.e., $R$ is 0.9. Considering the huge reduction in the computing costs of this setting, the loss of performance is still moderate and acceptable.

Fig.~\ref{fig:performance} shows the performance of the InversionNet across all the pruning ratio $R$ we have tried. Due to the limitation of the pages, we select 6 datasets out of the 10 datasets for demonstration. Note that here we use (1-SSIM) to replace SSIM such that a larger value of this metric implies a worse performance and thus is consistent with MAE and RMSE. It shows an increasing trend of these metrics when more filters in the InversionNet are pruned. And it clearly reflects a trade-off between computing costs and performance. Therefore, the user can choose an appropriate pruning ratio to get the lightweight InversionNet based on the requirement of their applications and the specifications of their edge devices.
\section{future direction}
Our work on the pruning of InversionNet is an early work to enable a machine learning-based method for FWI on edge devices. And there are several potential future directions to explore in this area. They are listed as follows.

\textbf{Unsupervised Learning}. The InversionNet used in this work is pre-trained and retrained under the supervised settings. However, the unsupervised settings are more practical for real applications since the measurements to get the labels of data are costly.~\cite{jin2021unsupervised} is the first work to explore effective unsupervised learning for FWI. However, how to develop an efficient and lightweight machine-learning model under unsupervised learning remained unexplored. Since unsupervised learning usually requires more diverse data than supervised learning, the customized data augmentation method may be necessary~\cite{wu2023synthetic, jin2023does, zhang2022toward, feng2023simplifying}. Moreover, how to design an effective pruning algorithm~\cite{ caron2020pruning} under unsupervised settings is also worth exploring.

\textbf{On-Device Learning}. Although our work has shown the potential of executing the inference on edge devices for FWI, more efforts still need to be made to implement the on-device learning for FWI, which can reduce the huge latency incurred by centralized training and address the potential privacy concerns~\cite{peng2023autorep, han2021deeprec, peng2023pasnet, xu2018deeptype, peng2023lingcn}. More specifically, the on-device training~\cite{wang2019e2, huang2023dynamic, luo2023efficient, wu2020enabling} requires more computing operations, storage space, and memory usage~\cite{bao2022accelerated, islam2019device, wang2021lightweight, bao2019efficient}, which is usually prohibitive for edge devices. Moreover, on-device learning is usually coupled with online learning~\cite{gu2023new, hayes2019memory, wu2021enabling} since the data collected on the device is often an online stream. Therefore, the collected data usually exhibits strong temporal correlations, which can not be ignored during training.

\textbf{Federated Learning}. In addition to the data augmentation mentioned above, another way to reduce the lack of labeled data for training is to apply federated learning to solve FWI problems~\cite{mcmahan2017communication, wu2021decentralized, li2020federated, wu2021federated, zhao2018federated, wu2022federated_3}. More specifically, the distributed devices can learn their local model from their local and small datasets. A global model is learned from the local models, which can perform well on the general dataset~\cite{wu2021federated_2,lim2020federated, li2020federated_2, wu2022distributed}.

\textbf{Deployment on Diverse Edge Devices}. In this work, Edge-InversionNet is mainly evaluated on the Raspberry Pi. However, there are multiple types of edge devices with their unique characteristics. When deployed on low-level microcontrollers~\cite{luo2023scaling, wu2020intermittent, zhang2022demo, lee2019intermittent, fedorov2019sparse, luo2019spoton}, more stringent hardware constraints should be considered. For the implementation of FPGAs~\cite{jiang2019accuracy, peng2022length, li2020edd}, the special circuits and hardware design should be fully exploited for speedup and energy savings. Besides, if the model is deployed on mobile GPUs~\cite{niu2020patdnn, xie2023accel, cai2021yolobile}, then how to parallelize the inference becomes much more important than the deployment on the other types of edge devices.

\section{Conclusions}
Among the data-driven machine learning models applied to seismic FWI,  InversionNet stands out as one of the most successful models. However, the considerable computing costs to run InversionNet have posed challenges in deploying it efficiently on edge devices, which often have limited computing resources.  To address this issue, we propose to utilize the structured pruning algorithm to generate a lightweight version of InversionNet that can perform efficient inference on edge devices. And we also developed a prototype to run the pruned InversionNet on a Raspberry Pi as a proof of concept. Experimental results show that structured pruning can reduce the size of InversionNet, save the computing resources to run it and speed up its execution on edge devices, without introducing a huge loss of performance.

\section*{ACKNOWLEDGMENTS}
This work was funded by the Los Alamos National Laboratory (LANL) - Technology Evaluation \& Demonstration (TED) Project. And This project was supported by resources provided by the Office of Research Computing at George Mason University (URL: https://orc.gmu.edu) and funded in part by grants from the National Science Foundation (Awards Number 1625039 and 2018631).

\bibliographystyle{unsrt}
\bibliography{reference,ref_lanl,ref_arxiv}

\begin{thebibliography}{10}

\bibitem{virieux2009overview}
Jean Virieux and St{\'e}phane Operto.
\newblock An overview of full-waveform inversion in exploration geophysics.
\newblock {\em Geophysics}, 74(6):WCC1--WCC26, 2009.

\bibitem{lin2023physics}
Youzuo Lin, James Theiler, and Brendt Wohlberg.
\newblock Physics-guided data-driven seismic inversion: Recent progress and future opportunities in full-waveform inversion.
\newblock {\em IEEE Signal Processing Magazine}, 40(1):115--133, 2023.

\bibitem{feng2022extremely}
Shihang Feng, Peng Jin, Xitong Zhang, Yinpeng Chen, David Alumbaugh, Michael Commer, and Youzuo Lin.
\newblock Extremely weak supervision inversion of multiphysical properties.
\newblock In {\em Second International Meeting for Applied Geoscience \& Energy}, pages 1785--1789. Society of Exploration Geophysicists and American Association of Petroleum~…, 2022.

\bibitem{wu2018deepdetect}
Yue Wu, Youzuo Lin, Zheng Zhou, David~Chas Bolton, Ji~Liu, and Paul Johnson.
\newblock Deepdetect: A cascaded region-based densely connected network for seismic event detection.
\newblock {\em IEEE Transactions on Geoscience and Remote Sensing}, 57(1):62--75, 2018.

\bibitem{lin2015quantifying}
Youzuo Lin and Lianjie Huang.
\newblock Quantifying subsurface geophysical properties changes using double-difference seismic-waveform inversion with a modified total-variation regularization scheme.
\newblock {\em Geophysical Supplements to the Monthly Notices of the Royal Astronomical Society}, 203(3):2125--2149, 2015.

\bibitem{feng2021multiscale}
Shihang Feng, Lei Fu, Zongcai Feng, and Gerard~T Schuster.
\newblock Multiscale phase inversion for vertical transverse isotropic media.
\newblock {\em Geophysical Prospecting}, 69(8-9):1634--1649, 2021.

\bibitem{yang2019deep}
Fangshu Yang and Jianwei Ma.
\newblock Deep-learning inversion: A next-generation seismic velocity model building method.
\newblock {\em Geophysics}, 84(4):R583--R599, 2019.

\bibitem{adler2020deep}
A~Adler, M~Araya-Polo, and T~Poggio.
\newblock Deep learning for seismic inverse problems: toward the acceleration of geophysical analysis workflows.
\newblock {\em IEEE Signal Processing Magazine}, 38(2):89--119, 2020.

\bibitem{feng2022intriguing}
Yinan Feng, Yinpeng Chen, Shihang Feng, Peng Jin, Zicheng Liu, and Youzuo Lin.
\newblock An intriguing property of geophysics inversion.
\newblock In {\em International Conference on Machine Learning}, pages 6434--6446. PMLR, 2022.

\bibitem{wu2019inversionnet}
Yue Wu and Youzuo Lin.
\newblock Inversionnet: An efficient and accurate data-driven full waveform inversion.
\newblock {\em IEEE Transactions on Computational Imaging}, 6:419--433, 2019.

\bibitem{zhang2020data}
Zhongping Zhang and Youzuo Lin.
\newblock Data-driven seismic waveform inversion: A study on the robustness and generalization.
\newblock {\em IEEE Transactions on Geoscience and Remote sensing}, 58(10):6900--6913, 2020.

\bibitem{zeng2021inversionnet3d}
Qili Zeng, Shihang Feng, Brendt Wohlberg, and Youzuo Lin.
\newblock Inversionnet3d: Efficient and scalable learning for 3-d full-waveform inversion.
\newblock {\em IEEE Transactions on Geoscience and Remote Sensing}, 60:1--16, 2021.

\bibitem{jin2022unsupervised}
Peng Jin, Xitong Zhang, Yinpeng Chen, Sharon~X Huang, Zicheng Liu, and Youzuo Lin.
\newblock Unsupervised learning of full-waveform inversion: Connecting {CNN} and partial differential equation in a loop.
\newblock In {\em International Conference on Learning Representations}, 2022.

\bibitem{manu2022seismic}
Daniel Manu, Petro~Mushidi Tshakwanda, Youzuo Lin, Weiwen Jiang, and Lei Yang.
\newblock Seismic waveform inversion capability on resource-constrained edge devices.
\newblock {\em Journal of Imaging}, 8(12):312, 2022.

\bibitem{li2017pruning}
Hao Li, Asim Kadav, Igor Durdanovic, Hanan Samet, and Hans~Peter Graf.
\newblock Pruning filters for efficient convnets.
\newblock In {\em International Conference on Learning Representations}, 2017.

\bibitem{liu2019metapruning}
Zechun Liu, Haoyuan Mu, Xiangyu Zhang, Zichao Guo, Xin Yang, Kwang-Ting Cheng, and Jian Sun.
\newblock Metapruning: Meta learning for automatic neural network channel pruning.
\newblock In {\em Proceedings of the IEEE/CVF international conference on computer vision}, pages 3296--3305, 2019.

\bibitem{li2022revisiting}
Yawei Li, Kamil Adamczewski, Wen Li, Shuhang Gu, Radu Timofte, and Luc Van~Gool.
\newblock Revisiting random channel pruning for neural network compression.
\newblock In {\em Proceedings of the IEEE/CVF Conference on Computer Vision and Pattern Recognition}, pages 191--201, 2022.

\bibitem{han2015deep_compression}
Song Han, Huizi Mao, and William~J Dally.
\newblock Deep compression: Compressing deep neural networks with pruning, trained quantization and huffman coding.
\newblock {\em International Conference on Learning Representations (ICLR)}, 2016.

\bibitem{peng2022towards}
Hongwu Peng, Deniz Gurevin, Shaoyi Huang, Tong Geng, Weiwen Jiang, Orner Khan, and Caiwen Ding.
\newblock Towards sparsification of graph neural networks.
\newblock In {\em 2022 IEEE 40th International Conference on Computer Design (ICCD)}, pages 272--279. IEEE, 2022.

\bibitem{wang2020sparsert}
Ziheng Wang.
\newblock Sparsert: Accelerating unstructured sparsity on gpus for deep learning inference.
\newblock In {\em Proceedings of the ACM international conference on parallel architectures and compilation techniques}, pages 31--42, 2020.

\bibitem{bao2020fast}
Runxue Bao, Bin Gu, and Heng Huang.
\newblock Fast oscar and owl regression via safe screening rules.
\newblock In {\em International Conference on Machine Learning}, pages 653--663. PMLR, 2020.

\bibitem{bao2022accelerated}
Runxue Bao, Bin Gu, and Heng Huang.
\newblock An accelerated doubly stochastic gradient method with faster explicit model identification.
\newblock In {\em Proceedings of the 31st ACM International Conference on Information \& Knowledge Management}, pages 57--66, 2022.

\bibitem{bao2022doubly}
Runxue Bao, Xidong Wu, Wenhan Xian, and Heng Huang.
\newblock Doubly sparse asynchronous learning for stochastic composite optimization.
\newblock In {\em Proceedings of the Thirty-First International Joint Conference on Artificial Intelligence, IJCAI}, pages 1916--1922, 2022.

\bibitem{deng2022openfwi}
Chengyuan Deng, Shihang Feng, Hanchen Wang, Xitong Zhang, Peng Jin, Yinan Feng, Qili Zeng, Yinpeng Chen, and Youzuo Lin.
\newblock Openfwi: Large-scale multi-structural benchmark datasets for full waveform inversion.
\newblock {\em Advances in Neural Information Processing Systems}, 35:6007--6020, 2022.

\bibitem{jin2021unsupervised}
Peng Jin, Xitong Zhang, Yinpeng Chen, Sharon~Xiaolei Huang, Zicheng Liu, and Youzuo Lin.
\newblock Unsupervised learning of full-waveform inversion: Connecting cnn and partial differential equation in a loop.
\newblock {\em arXiv preprint arXiv:2110.07584}, 2021.

\bibitem{wu2023synthetic}
Yawen Wu, Zhepeng Wang, Dewen Zeng, Yiyu Shi, and Jingtong Hu.
\newblock Synthetic data can also teach: Synthesizing effective data for unsupervised visual representation learning.
\newblock In {\em Proceedings of the AAAI Conference on Artificial Intelligence}, volume~37, pages 2866--2874, 2023.

\bibitem{jin2023does}
Peng Jin, Yinan Feng, Shihang Feng, Hanchen Wang, Yinpeng Chen, Benjamin Consolvo, Zicheng Liu, and Youzuo Lin.
\newblock Does full waveform inversion benefit from big data?
\newblock {\em arXiv preprint arXiv:2307.15388}, 2023.

\bibitem{zhang2022toward}
Yanfu Zhang, Runxue Bao, Jian Pei, and Heng Huang.
\newblock Toward unified data and algorithm fairness via adversarial data augmentation and adaptive model fine-tuning.
\newblock In {\em 2022 IEEE International Conference on Data Mining (ICDM)}, pages 1317--1322. IEEE, 2022.

\bibitem{feng2023simplifying}
Yinan Feng, Yinpeng Chen, Peng Jin, Shihang Feng, Zicheng Liu, and Youzuo Lin.
\newblock Simplifying full waveform inversion via domain-independent self-supervised learning.
\newblock {\em arXiv preprint arXiv:2305.13314}, 2023.

\bibitem{caron2020pruning}
Mathilde Caron, Ari Morcos, Piotr Bojanowski, Julien Mairal, and Armand Joulin.
\newblock Pruning convolutional neural networks with self-supervision.
\newblock {\em arXiv preprint arXiv:2001.03554}, 2020.

\bibitem{peng2023autorep}
Hongwu Peng, Shaoyi Huang, Tong Zhou, Yukui Luo, Chenghong Wang, Zigeng Wang, Jiahui Zhao, Xi~Xie, Ang Li, Tony Geng, et~al.
\newblock Autorep: Automatic relu replacement for fast private network inference.
\newblock In {\em Proceedings of the IEEE/CVF International Conference on Computer Vision}, pages 5178--5188, 2023.

\bibitem{han2021deeprec}
Jialiang Han, Yun Ma, Qiaozhu Mei, and Xuanzhe Liu.
\newblock Deeprec: On-device deep learning for privacy-preserving sequential recommendation in mobile commerce.
\newblock In {\em Proceedings of the Web Conference 2021}, pages 900--911, 2021.

\bibitem{peng2023pasnet}
Hongwu Peng, Shanglin Zhou, Yukui Luo, Nuo Xu, Shijin Duan, Ran Ran, Jiahui Zhao, Chenghong Wang, Tong Geng, Wujie Wen, et~al.
\newblock Pasnet: Polynomial architecture search framework for two-party computation-based secure neural network deployment.
\newblock In {\em 2023 60th ACM/IEEE Design Automation Conference (DAC)}, pages 1--6. IEEE, 2023.

\bibitem{xu2018deeptype}
Mengwei Xu, Feng Qian, Qiaozhu Mei, Kang Huang, and Xuanzhe Liu.
\newblock Deeptype: On-device deep learning for input personalization service with minimal privacy concern.
\newblock {\em Proceedings of the ACM on Interactive, Mobile, Wearable and Ubiquitous Technologies}, 2(4):1--26, 2018.

\bibitem{peng2023lingcn}
Hongwu Peng, Ran Ran, Yukui Luo, Jiahui Zhao, Shaoyi Huang, Kiran Thorat, Tong Geng, Chenghong Wang, Xiaolin Xu, Wujie Wen, et~al.
\newblock Lingcn: Structural linearized graph convolutional network for homomorphically encrypted inference.
\newblock {\em arXiv preprint arXiv:2309.14331}, 2023.

\bibitem{wang2019e2}
Yue Wang, Ziyu Jiang, Xiaohan Chen, Pengfei Xu, Yang Zhao, Yingyan Lin, and Zhangyang Wang.
\newblock E2-train: Training state-of-the-art cnns with over 80\% energy savings.
\newblock {\em Advances in Neural Information Processing Systems}, 32, 2019.

\bibitem{huang2023dynamic}
Shaoyi Huang, Bowen Lei, Dongkuan Xu, Hongwu Peng, Yue Sun, Mimi Xie, and Caiwen Ding.
\newblock Dynamic sparse training via balancing the exploration-exploitation trade-off.
\newblock In {\em 2023 60th ACM/IEEE Design Automation Conference (DAC)}, pages 1--6. IEEE, 2023.

\bibitem{luo2023efficient}
Yubo Luo, Le~Zhang, Zhenyu Wang, and Shahriar Nirjon.
\newblock Efficient multitask learning on resource-constrained systems.
\newblock {\em arXiv preprint arXiv:2302.13155}, 2023.

\bibitem{wu2020enabling}
Yawen Wu, Zhepeng Wang, Yiyu Shi, and Jingtong Hu.
\newblock Enabling on-device cnn training by self-supervised instance filtering and error map pruning.
\newblock {\em IEEE Transactions on Computer-Aided Design of Integrated Circuits and Systems}, 39(11):3445--3457, 2020.

\bibitem{islam2019device}
Bahsima Islam, Yubo Luo, Seulki Lee, and Shahriar Nirjon.
\newblock On-device training from sensor data on batteryless platforms.
\newblock In {\em Proceedings of the 18th International Conference on Information Processing in Sensor Networks}, pages 325--326, 2019.

\bibitem{wang2021lightweight}
Zhepeng Wang, Yawen Wu, Zhenge Jia, Yiyu Shi, and Jingtong Hu.
\newblock Lightweight run-time working memory compression for deployment of deep neural networks on resource-constrained mcus.
\newblock In {\em Proceedings of the 26th Asia and South Pacific Design Automation Conference}, pages 607--614, 2021.

\bibitem{bao2019efficient}
Runxue Bao, Bin Gu, and Heng Huang.
\newblock Efficient approximate solution path algorithm for order weight l\_1-norm with accuracy guarantee.
\newblock In {\em 2019 IEEE International Conference on Data Mining (ICDM)}, pages 958--963. IEEE, 2019.

\bibitem{gu2023new}
Bin Gu, Runxue Bao, Chenkang Zhang, and Heng Huang.
\newblock New scalable and efficient online pairwise learning algorithm.
\newblock {\em IEEE Transactions on Neural Networks and Learning Systems}, 2023.

\bibitem{hayes2019memory}
Tyler~L Hayes, Nathan~D Cahill, and Christopher Kanan.
\newblock Memory efficient experience replay for streaming learning.
\newblock In {\em 2019 International Conference on Robotics and Automation (ICRA)}, pages 9769--9776. IEEE, 2019.

\bibitem{wu2021enabling}
Yawen Wu, Zhepeng Wang, Dewen Zeng, Yiyu Shi, and Jingtong Hu.
\newblock Enabling on-device self-supervised contrastive learning with selective data contrast.
\newblock In {\em 2021 58th ACM/IEEE Design Automation Conference (DAC)}, pages 655--660. IEEE, 2021.

\bibitem{mcmahan2017communication}
Brendan McMahan, Eider Moore, Daniel Ramage, Seth Hampson, and Blaise~Aguera y~Arcas.
\newblock Communication-efficient learning of deep networks from decentralized data.
\newblock In {\em Artificial intelligence and statistics}, pages 1273--1282. PMLR, 2017.

\bibitem{wu2021decentralized}
Yawen Wu, Zhepeng Wang, Dewen Zeng, Meng Li, Yiyu Shi, and Jingtong Hu.
\newblock Decentralized unsupervised learning of visual representations.
\newblock {\em arXiv preprint arXiv:2111.10763}, 2021.

\bibitem{li2020federated}
Tian Li, Anit~Kumar Sahu, Manzil Zaheer, Maziar Sanjabi, Ameet Talwalkar, and Virginia Smith.
\newblock Federated optimization in heterogeneous networks.
\newblock {\em Proceedings of Machine learning and systems}, 2:429--450, 2020.

\bibitem{wu2021federated}
Yawen Wu, Dewen Zeng, Zhepeng Wang, Yi~Sheng, Lei Yang, Alaina~J James, Yiyu Shi, and Jingtong Hu.
\newblock Federated contrastive learning for dermatological disease diagnosis via on-device learning.
\newblock In {\em 2021 IEEE/ACM International Conference On Computer Aided Design (ICCAD)}, pages 1--7. IEEE, 2021.

\bibitem{zhao2018federated}
Yue Zhao, Meng Li, Liangzhen Lai, Naveen Suda, Damon Civin, and Vikas Chandra.
\newblock Federated learning with non-iid data.
\newblock {\em arXiv preprint arXiv:1806.00582}, 2018.

\bibitem{wu2022federated_3}
Yawen Wu, Dewen Zeng, Zhepeng Wang, Yi~Sheng, Lei Yang, Alaina~J James, Yiyu Shi, and Jingtong Hu.
\newblock Federated self-supervised contrastive learning and masked autoencoder for dermatological disease diagnosis.
\newblock {\em arXiv preprint arXiv:2208.11278}, 2022.

\bibitem{wu2021federated_2}
Yawen Wu, Dewen Zeng, Zhepeng Wang, Yiyu Shi, and Jingtong Hu.
\newblock Federated contrastive learning for volumetric medical image segmentation.
\newblock In {\em Medical Image Computing and Computer Assisted Intervention--MICCAI 2021: 24th International Conference, Strasbourg, France, September 27--October 1, 2021, Proceedings, Part III 24}, pages 367--377. Springer, 2021.

\bibitem{lim2020federated}
Wei Yang~Bryan Lim, Nguyen~Cong Luong, Dinh~Thai Hoang, Yutao Jiao, Ying-Chang Liang, Qiang Yang, Dusit Niyato, and Chunyan Miao.
\newblock Federated learning in mobile edge networks: A comprehensive survey.
\newblock {\em IEEE Communications Surveys \& Tutorials}, 22(3):2031--2063, 2020.

\bibitem{li2020federated_2}
Tian Li, Anit~Kumar Sahu, Ameet Talwalkar, and Virginia Smith.
\newblock Federated learning: Challenges, methods, and future directions.
\newblock {\em IEEE signal processing magazine}, 37(3):50--60, 2020.

\bibitem{wu2022distributed}
Yawen Wu, Dewen Zeng, Zhepeng Wang, Yiyu Shi, and Jingtong Hu.
\newblock Distributed contrastive learning for medical image segmentation.
\newblock {\em Medical Image Analysis}, 81:102564, 2022.

\bibitem{luo2023scaling}
Yubo Luo.
\newblock {\em Scaling Up Task Execution on Resource-Constrained Systems}.
\newblock PhD thesis, The University of North Carolina at Chapel Hill, 2023.

\bibitem{wu2020intermittent}
Yawen Wu, Zhepeng Wang, Zhenge Jia, Yiyu Shi, and Jingtong Hu.
\newblock Intermittent inference with nonuniformly compressed multi-exit neural network for energy harvesting powered devices.
\newblock In {\em 2020 57th ACM/IEEE Design Automation Conference (DAC)}, pages 1--6. IEEE, 2020.

\bibitem{zhang2022demo}
Le~Zhang, Yubo Luo, and Shahriar Nirjon.
\newblock Demo abstract: Capuchin: A neural network model generator for 16-bit microcontrollers.
\newblock In {\em 2022 21st ACM/IEEE International Conference on Information Processing in Sensor Networks (IPSN)}, pages 497--498. IEEE, 2022.

\bibitem{lee2019intermittent}
Seulki Lee, Bashima Islam, Yubo Luo, and Shahriar Nirjon.
\newblock Intermittent learning: On-device machine learning on intermittently powered system.
\newblock {\em Proceedings of the ACM on Interactive, Mobile, Wearable and Ubiquitous Technologies}, 3(4):1--30, 2019.

\bibitem{fedorov2019sparse}
Igor Fedorov, Ryan~P Adams, Matthew Mattina, and Paul Whatmough.
\newblock Sparse: Sparse architecture search for cnns on resource-constrained microcontrollers.
\newblock {\em Advances in Neural Information Processing Systems}, 32, 2019.

\bibitem{luo2019spoton}
Yubo Luo and Shahriar Nirjon.
\newblock Spoton: Just-in-time active event detection on energy autonomous sensing systems.
\newblock {\em Brief Presentations Proceedings (RTAS 2019)}, 9, 2019.

\bibitem{jiang2019accuracy}
Weiwen Jiang, Xinyi Zhang, Edwin H-M Sha, Lei Yang, Qingfeng Zhuge, Yiyu Shi, and Jingtong Hu.
\newblock Accuracy vs. efficiency: Achieving both through fpga-implementation aware neural architecture search.
\newblock In {\em Proceedings of the 56th Annual Design Automation Conference 2019}, pages 1--6, 2019.

\bibitem{peng2022length}
Hongwu Peng, Shaoyi Huang, Shiyang Chen, Bingbing Li, Tong Geng, Ang Li, Weiwen Jiang, Wujie Wen, Jinbo Bi, Hang Liu, et~al.
\newblock A length adaptive algorithm-hardware co-design of transformer on fpga through sparse attention and dynamic pipelining.
\newblock In {\em Proceedings of the 59th ACM/IEEE Design Automation Conference}, pages 1135--1140, 2022.

\bibitem{li2020edd}
Yuhong Li, Cong Hao, Xiaofan Zhang, Xinheng Liu, Yao Chen, Jinjun Xiong, Wen-mei Hwu, and Deming Chen.
\newblock Edd: Efficient differentiable dnn architecture and implementation co-search for embedded ai solutions.
\newblock In {\em 2020 57th ACM/IEEE Design Automation Conference (DAC)}, pages 1--6. IEEE, 2020.

\bibitem{niu2020patdnn}
Wei Niu, Xiaolong Ma, Sheng Lin, Shihao Wang, Xuehai Qian, Xue Lin, Yanzhi Wang, and Bin Ren.
\newblock Patdnn: Achieving real-time dnn execution on mobile devices with pattern-based weight pruning.
\newblock In {\em Proceedings of the Twenty-Fifth International Conference on Architectural Support for Programming Languages and Operating Systems}, pages 907--922, 2020.

\bibitem{xie2023accel}
Xi~Xie, Hongwu Peng, Amit Hasan, Shaoyi Huang, Jiahui Zhao, Haowen Fang, Wei Zhang, Tong Geng, Omer Khan, and Caiwen Ding.
\newblock Accel-gcn: High-performance gpu accelerator design for graph convolution networks.
\newblock {\em arXiv preprint arXiv:2308.11825}, 2023.

\bibitem{cai2021yolobile}
Yuxuan Cai, Hongjia Li, Geng Yuan, Wei Niu, Yanyu Li, Xulong Tang, Bin Ren, and Yanzhi Wang.
\newblock Yolobile: Real-time object detection on mobile devices via compression-compilation co-design.
\newblock In {\em Proceedings of the AAAI conference on artificial intelligence}, volume~35, pages 955--963, 2021.

\end{thebibliography}
\end{document}